\documentclass[10pt,twocolumn,letterpaper]{article}

\usepackage{cvpr}              

\usepackage[dvipsnames]{xcolor}

\definecolor{cvprblue}{rgb}{0.21,0.49,0.74}
\usepackage[pagebackref,breaklinks,colorlinks,citecolor=cvprblue]{hyperref}
\usepackage{multirow}
\usepackage{times}
\usepackage{epsfig}
\usepackage{graphicx}
\usepackage{amsmath}
\usepackage{amssymb}
\usepackage{booktabs}
\usepackage{makecell}
\usepackage{multirow}
\usepackage{caption}
\usepackage{gensymb}
\usepackage[lined,boxed,commentsnumbered,ruled]{algorithm2e}
\usepackage{rotating}
\usepackage{bm}
\usepackage{colortbl}
\usepackage{color}
\usepackage{pifont}
\usepackage{makecell} 
\usepackage{wrapfig}
\usepackage{xspace}
\usepackage{hhline}
\usepackage{csquotes}

\usepackage[accsupp]{axessibility}

\def\paperID{1641} 
\def\confName{CVPR}
\def\confYear{2024}

\title{PointDGRWKV: Generalizing RWKV-like Architecture to Unseen Domains \\for Point Cloud Classification}

\author{Hao Yang$^{1}$\footnotemark[1], 
Qianyu Zhou$^{2}$\thanks{\textit{The first two authors contributed equally to this work.}}, 
Haijia Sun$^3$,
Xiangtai Li$^4$, 
Xuequan Lu$^5$, \\
Lizhuang Ma$^1$\thanks{\textit{Corresponding author.}},
Shuicheng Yan$^{6}$
\\$^1$Shanghai Jiao Tong University; 
$^2$ The University of Tokyo;
$^3$ Nanjing University;  \\
$^4$ Nanyang Technological University;
$^5$  The University of Western Australia;
$^6$ National University of Singapore
\\
{\tt\small \textbf{Code:}\quad \url{https://github.com/yxltya/PointDGRWKV}}
}

\begin{document}

\maketitle

\begin{abstract}
    Domain Generalization (DG) has been recently explored to enhance the generalizability of Point Cloud Classification (PCC) models toward unseen domains. Prior works are based on convolutional networks, Transformer or Mamba architectures, either suffering from limited receptive fields or high computational cost, or insufficient long-range dependency modeling. 
    RWKV, as an emerging architecture, possesses superior linear complexity, global receptive fields, and long-range dependency. In this paper, we present the first work that studies the generalizability of RWKV models in DG PCC. We find that directly applying RWKV to DG PCC encounters \textit{two significant challenges}: RWKV's fixed direction token shift methods, like Q-Shift, introduce spatial distortions when applied to unstructured point clouds, weakening local geometric modeling and reducing robustness. In addition, the Bi-WKV attention in RWKV amplifies slight cross-domain differences in \textit{key} distributions through exponential weighting, leading to attention shifts and degraded generalization. 
    To this end, we propose PointDGRWKV, the first RWKV-based framework tailored for DG PCC. It introduces two key modules to enhance spatial modeling and cross-domain robustness, while maintaining RWKV's linear efficiency. In particular, we present Adaptive Geometric Token Shift to model local neighborhood structures to improve geometric context awareness. In addition, Cross-Domain \textit{key} feature Distribution Alignment is designed to mitigate attention drift by aligning \textit{key} feature distributions across domains. Extensive experiments on multiple benchmarks demonstrate that PointDGRWKV achieves state-of-the-art performance on DG PCC.  
\end{abstract}

\section{Introduction}

3D point clouds analysis~\cite{hackel2017semantic3d,uy2019revisiting,ren2022benchmarking,ben20183dmfv,li2020pointaugment,zhang2018graph,qiu2021dense,qiu2021geometric,zhang2020pointhop,zhang2023deep,xu2020geometry} play a crucial role in various applications, such as autonomous driving, augmented reality, and robotics~\cite{caesar2020nuscenes, billinghurst2015survey, thuruthel2019soft}. 
Recently, point cloud classification (PCC) tasks~\cite{qi2017pointnet,qi2017pointnet++,phan2018dgcnn,li2018pointcnn,wang2019dynamic,ben20183dmfv,zhang2018graph} have made significant progress in understanding the local geometry and global shapes. 
However, most of them usually assume that the training and testing data share the same distribution. When the model is applied to unknown domains, the performance often drops significantly due to domain shifts induced by different sensors, environments,  scanning angles, \emph{etc}.

\begin{figure}[t]
\centering
\label{fig:teaser}
\includegraphics[scale=0.30]{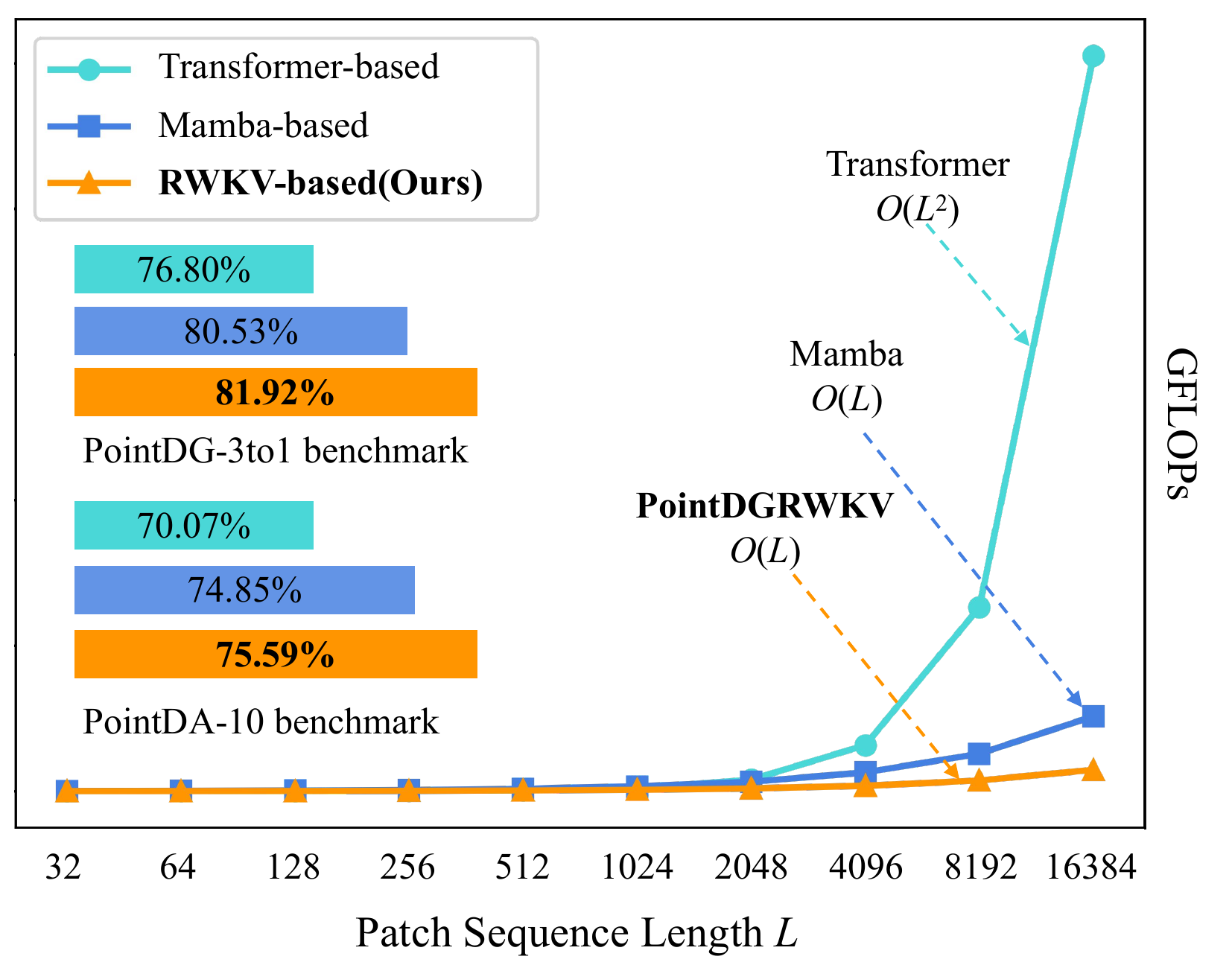}
\caption{
Accuracy-speed tradeoff in DG PCC. 
(Left) Overall accuracy of methods with different architectures (Right)
FLOPs increase with sequence length. Our PointDGRWKV achieves superior performance with its linear complexity.
}
\label{teaser}
\end{figure}

To address this issue, domain generalization (DG) has been introduced into point cloud analysis,  aiming to train models solely on source domains and generalize well in unknown domains. The mainstream DG PCC methods tend to learn domain-invariant features via data augmentation~\cite{xiao20233d}, adversarial training~\cite{lehner20223d}, and consistency learning~\cite{kim2023single}. 
Nonetheless, most of them are based on CNNs, 
and suffer from a limited receptive field, making it challenging to capture global structural information and harm the generalizability. Subsequently, point Transformer~\cite{han2022survey} was introduced to enhance global modeling capabilities in DG PCC.
However, its inherent attention involves high computational complexity, which limits its efficiency in practical applications. 
Point Cloud Mamba has recently shown the potential of sequence modeling in DG PCC~\cite{yang2025pointdgmamba}. Nevertheless, due to its fixed state space size, it is difficult to fully capture long-range dependencies, especially under long sequence lengths, as shown in Figure~\ref{teaser}.

Recently, Reception Weighted Key Value (RWKV)~\cite{peng2023rwkv,chen2025zig,duan2024vision,dai2025stylerwkv} has demonstrated excellent capabilities in long-range dependency modeling and capturing global information in NLP and vision tasks. 
Moreover, the core WKV attention mechanism exhibits a linear computational complexity, which significantly reduces the computational overhead of traditional self-attention. 
They demonstrate strong scalability in various vision tasks and even in point cloud analysis. 
Despite its gratifying progress, enhancing the generalizability of RWKV-like models in unseen domains for point cloud analysis remains an open problem, as directly applying RWKV to DG PCC tasks is non-trivial.

In this paper, we aim to improve the generalizability of RWKV-like architectures toward unseen domains in point cloud classification. 
Our motivations mainly lie in \textit{two} aspects. \textit{Firstly}, RWKV's fixed direction token shift, \emph{e.g.,} Q-Shift, would inevitably introduce spatial distortions to unstructured point clouds due to the inconsistent order of token arrangement and spatial proximity, weakening the model's ability to model local geometry and thus affecting robustness in unseen domains. 
Secondly, the Bi-WKV attention mechanism in RWKV is highly sensitive to slight discrepancies in \textit{key} distribution between the source domain and the unseen domain. The nonlinear amplification characteristics, \emph{i.e.,} the exponential function, can easily amplify the shift in the focus of attention, undermining the generalization performance of the model in unknown domains.

Motivated by the aforementioned analysis, we propose PointDGRWKV, a novel RWKV-based framework for domain generalized point cloud classification. 
PointDGRWKV excels in strong generalizability, linear complexity, and capabilities in modeling long-range dependency and global structure information. 
Our proposed method has two key modules. 
\textbf{Firstly}, we design a lightweight, parameter-free Adaptive Geometric Token Shift mechanism (AGT-Shift) based on the inherent spatial characteristics of point clouds. It constructs local neighborhoods through spatial partitioning and dynamically integrates structural features to enhance the model's ability to model geometric contexts. This mechanism is specifically designed based on the characteristics of point clouds.
\textbf{Secondly}, we propose a Cross-Domain \textit{Key} feature Distribution Alignment module (CD-KDA) to address the nonlinear amplification effect of \textit{key} vectors on weight calculation in the Bi-WKV attention mechanism. 
By aligning the \textit{key} distributions between source domains at the mean and covariance levels, we explicitly alleviate the cross-domain shift of attention and improve the generalization performance of the model in unseen domains. 
As shown in Fig.~\ref{teaser}, PointDGRWKV achieves superior performance with less computational overhead compared to existing Transformer-based and Mamba-based methods on multiple DG benchmarks.
Our contributions are three-fold:

\begin{itemize}
    \item We propose PointDGRWKV, a novel RWKV-based framework for domain generalizable point cloud classification that excels in strong generalizability toward unseen domains, global receptive fields, linear complexity, and long-range dependency.
    \item We design Adaptive Geometric Token Shift (AGT-Shift) and Cross-Domain \textit{key} feature Distribution Alignment (CD-KDA) to enhance RWKV's 
    geometry perception ability and the generalizability toward unseen domains. 
    \item Extensive experiments on multiple DG benchmarks verify the superiority and effectiveness of PointDGRWKV compared to state-of-the-art approaches. 
\end{itemize}

\section{Related Work}
\noindent {\bf{Point Cloud Classification}} (PCC) aims to accurately categorize 3D point cloud data. Early works such as PointNet~\cite{qi2017pointnet} and PointNet++\cite{qi2017pointnet++} pioneered the use of MLP-based architectures to directly learn features from raw point clouds. Subsequent research expanded on this by incorporating Convolutional Neural Networks (CNNs)\cite{li2018pointcnn,wang2019dynamic} to better capture local geometric patterns.
Nevertheless, CNN-based approaches often struggle with limited receptive fields, especially in deeper networks. To address this, Vision Transformers (ViTs)\cite{zhao2021point,fang2024explore,deng2024vg4d} have recently been adopted in PCC, offering enhanced global context modeling capabilities. Methods like PCT\cite{guo2021pct} and Point Transformer~\cite{zhao2021point} leverage self-attention mechanisms to capture long-range dependencies across points. 
Recently, Point Mamba and Point Cloud Mamba~\cite{liang2024pointmamba,zhang2025point} have introduced Mamba-like models into point cloud analysis, and achieved a global receptive field with linear complexity.
While these models achieve impressive results on standard benchmarks, their generalization to novel or unseen domains remains a significant challenge.

\noindent {\bf{Domain Generalized Point Cloud Classification}} (DG PCC) Although domain adaptation techniques~\cite{ganin2015unsupervised, wang2018deep,zhou2022generative,zhou2023context,zhou2023self,gu2021pit,zhou2022uncertainty,xu2021semi,feng2022dmt,zhou2022domainb,guo2021label,zhou2022uncertainty} have been explored in point cloud areas~\cite{qin2019pointdan,zou2021geometry,wang2021cross,shen2022domain,wang2024unsupervised,liang2022point,fan2022self,huang2022generation,wu2019squeezesegv2,katageri2024synergizing,katageri2024synergizing,liu2024cloudmix,jiang2024pcotta} to narrow the domain shifts, the target data is not always accessible in real scenarios, which might fail these methods. Domain generalization~\cite{wang2022generalizing,zhou2022domain,long2024dgmamba,jiang2024dgpic,wang2024disentangle,long2024rethink,zhou2024test,zhou2023instance,zhou2022adaptive,song2024susam,long2025diverse,song2024ba,long2025domain} has recently been introduced into PCC~\cite{xu2024push,kim2023single,lehner20223d,xiao20233d,xiao2022learning,huang2023sug,huang2021metasets,wei2022learning} to improve the generalizability toward unseen domains.  Existing DG PCC methods primarily focus on learning domain-invariant representations through meta learning~\cite {huang2021metasets}, adversarial learning~\cite{xu2024push}, contrastive learning~\cite{wei2022learning}
consistency regularization~\cite{kim2023single}, and data augmentation~\cite{lehner20223d,xiao20233d,xiao2022learning}. While these approaches have shown promising results, many are built on CNN-based backbones, whose inherently limited receptive fields constrain their ability to capture global structural information critical for robust generalization. Subsequently, Huang et al.~\cite{huang2023sug} proposed Transformers-based subdomain alignment and domain-aware attention mechanisms, while suffer from the quadratic computational costs. Recently, PointDGMamba~\cite{yang2025pointdgmamba} introduced Mamba-based architectures in DG PCC to improve generalization to unseen domains. Although Mamba offers linear inference efficiency and long-sequence modeling capabilities, its fixed-size state space constrains its ability to capture long-range spatial dependencies in point clouds. This highlights the need for novel architectures in DG PCC that can simultaneously support global context modeling and maintain better long-range dependencies on long sequence length.

\begin{figure}[t]
\centering
\vspace{2mm}
\includegraphics[scale=0.78]{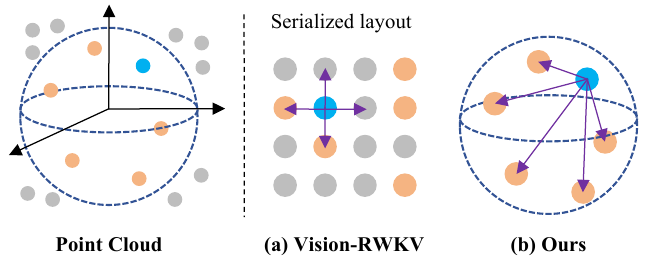}
\caption{
Comparison of local structure modeling between Vision-RWKV and our PointDGRWKV. 
(a) Vision-RWKV uses fixed four-directional token shifts, which may distort spatial relationships in unstructured point clouds. (b) In contrast, our Adaptive Geometric Token Shift dynamically aggregates local information based on geometric neighborhood, making it more suitable for point cloud data.}
\label{3.2D_3D_receptive_field}
\end{figure}

\noindent {\bf{Reception Weighted Key Value (RWKV).}} RWKV~\cite{yuan2024mamba,chen2025zig,yin2024video,dai2025stylerwkv} has garnered increasing attention due to its significant advantages in global receptive fields, computational complexity, and advantages in long sequence modeling. The core innovation is its linear WKV attention mechanism and spatial mixing and channel mixing, balancing local features and global dependencies through gating and recursion mechanisms, and supporting parallel training and efficient inference. Recently, Point-RWKV~\cite{he2025pointrwkv} introduced RWKV in point cloud analysis, but did not really open-source their implementations.  Regarding the unstructured and sparse nature of point clouds, as well as cross-domain differences such as sensor or scene changes, pose new challenges to the original RWKV.
To our knowledge, this is the first work that studies the generalizability of RWKV-based models toward unseen domains in point cloud tasks. This paper uses the popular vision-RWKV~\cite{yuan2024mamba} as the baseline.

\section{Method}
\subsection{Revisiting the RWKV}

\begin{figure}[t]
\centering
\vspace{4.5mm}
\includegraphics[scale=0.85]{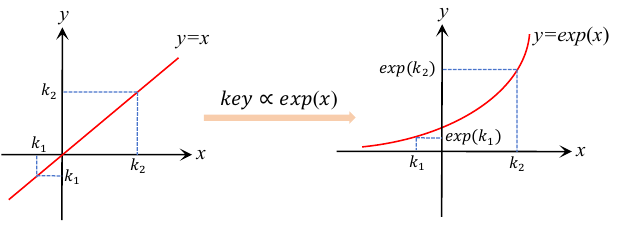}

\caption{
Illustration of the exponential function's amplification effect on \textit{key} differences: for example, $k_1 \text{ = - 0.3}$ and $k_2 \text{ = 1.0}$ differ by only 1.3, but after exponentiation, they become $e^{-0.3} \approx \text{0.74}$ and $e^{1.0} \approx \text{2.72}$, showing a magnified gap. This indicates that in Bi-WKV attention, small \textit{key} differences can be significantly amplified, potentially causing biased attention and harming the model's generalizability.}
\label{4.domain_1_2_key}
\end{figure}

\begin{figure*}[!ht]
\centering
\includegraphics[scale=0.60]{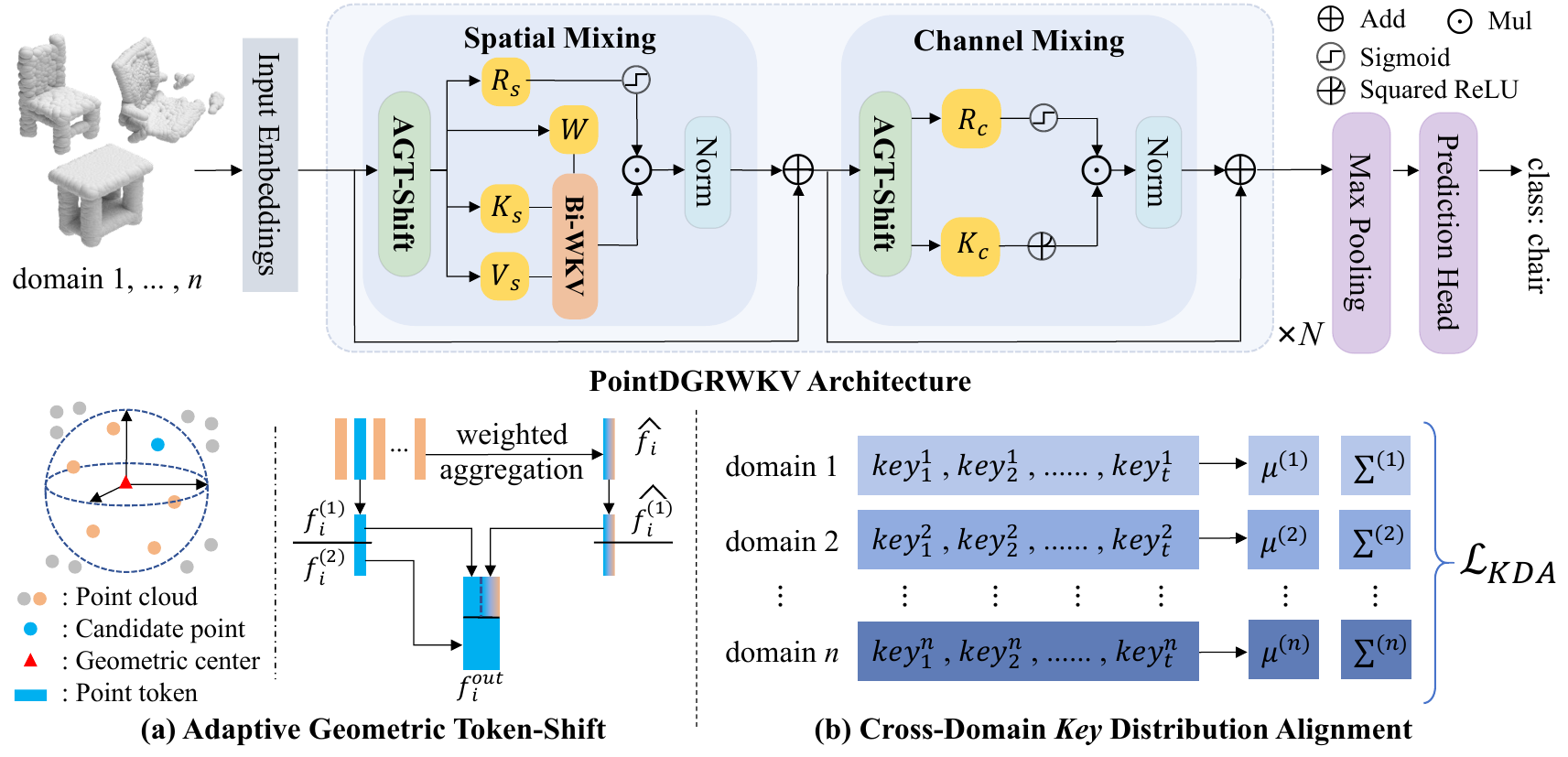}
\caption{The architecture of PointDGRWKV consists of two key components:
(a) \textbf{Adaptive Geometric Token-Shift}: it introduces an adaptive offset strategy based on geometric relations to fuse features from neighboring points in a weighted manner. It enables structure-aware feature shift and enhancement, improving the model's ability to capture local geometric structures.
(b) \textbf{Cross-Domain \textit{Key} Distribution Alignment}: By minimizing the distribution differences of \textit{key} across different source domains, CD-KDA reduces inconsistencies in cross-domain attention and enhances the model’s generalizability across domains.}
\vspace{7mm}
\label{1-network}
\end{figure*}

RWKV~\cite{duan2024vision} incorporates a token shift function, \emph{e.g.,} Q-Shift, which introduces interactions among nearby positions along the channel dimension, enriching local context without increasing the computational cost:

\begin{equation}
\begin{aligned}
\label{eq-Q-shift}
    \text{Q-Shift}_{\mathcal{S}}(X) &= X + (1 - \mu_{\mathcal{S}}) X^{\star} \\
    X^\star &= \text{Concat}(X_1, X_2, X_3, X_4),
\end{aligned}
\end{equation}

\noindent where $X^\star$ denotes a sliced vector of $X$, capturing tokens from positions adjacent to the current location in the channel dimension. However, when directly applied to unstructured 3D point clouds, this operation may distort the underlying spatial structure (Fig.\ref{3.2D_3D_receptive_field}).

Moreover, in the attention mechanism adopted by $\text{Bi-WKV}$~\cite{duan2024vision}, the attention weight of each token is formulated as follows:
\begin{equation}
\begin{aligned}
\label{eq-a-3}
\mathrm{wkv}_t &= \text{Bi-WKV}(K,V)_t \\
&= {\frac{
\sum_{i=0,i \ne t}^{T-1} e^{-\frac{|t - i| - 1}{T} \cdot \mathbf{w} + \mathbf{k}_i} \cdot \mathbf{v}_i + e^{\mathbf{u} + \mathbf{k}_t} \cdot \mathbf{v}_t
}{
\sum_{i=0, i \ne t}^{T-1} e^{-\frac{|t - i| - 1}{T} \cdot \mathbf{w} + \mathbf{k}_i} + e^{\mathbf{u} + \mathbf{k}_t}} ,
}
\end{aligned}
\end{equation}
where $\mathbf{k}_i$ and $\mathbf{v}_i$ represent the \textit{key} and \textit{value} of the $\mathbf{i}\text{-}th$ token, respectively, and $\mathbf{w}$ is the learnable distance decay parameter, and $\mathbf{u}$ is the learnable bias term. Since the \textit{key} $\mathbf{k}$ appears directly in the exponential function, its distribution has an exponential amplification effect on attention results (Fig.\ref{4.domain_1_2_key}), which can lead to attention drift and degrade the model's generalization performance.

To address these limitations in the context of point cloud domain generalization, we propose two modules, as illustrated in Fig.~\ref{1-network}: Adaptive Geometric Token Shift (AGT-Shift), which enhances local structure modeling via spatial partitioning, and Cross-Domain \textit{Key} Distribution Alignment (CD-KDA), which improves the robustness of attention by aligning \textit{key} feature distributions across domains.

\subsection{Adaptive Geometric Token-Shift}
When adapting the token shift mechanism of RWKV to the point cloud domain, two key challenges arise. 
Firstly, point clouds inherently lack regular topological structures, making it difficult to establish consistent spatial directions such as ``up," ``down," ``left," and ``right" as in image data. 
Secondly, point cloud datasets are typically large-scale, and conventional operations like KNN search or graph construction introduce substantial computational and memory overhead, thereby limiting scalability. 
Consequently, a central challenge lies in achieving an effective balance between computational efficiency and the ability to model spatial structures.

To address this issue, we propose Adaptive Geometric Token Shift (AGT-Shift). AGT-Shift efficiently constructs the nearest neighborhood through spatial partitioning and introduces a weighted feature aggregation scheme among neighboring points to enable token shifting and enhance structural awareness. By avoiding the explicit computation of pairwise distance matrices, the method circumvents the quadratic complexity typically found in KNN-based approaches.

Concretely, AGT-Shift partitions the 3D space into a set of spatial sub-regions using a spatial hashing technique with fixed step sizes. Points within the same sub-region are treated as forming a local context block, capturing localized geometric structures. For each point token, partial feature fusion is then conducted by computing a weighted average of the feature tokens within its corresponding region, facilitating efficient context-aware feature enhancement.

Let the point cloud features be denoted as $F \in \mathbb{R}^{B \times N \times C}$  and the corresponding point coordinates as $X \in \mathbb{R}^{B \times N \times 3}$, where $B$, $N$, $C$, represent the batch size, number of points, and feature dimensions, respectively. The 3D space is discretized into a set of spatial grids $\mathcal{G}_i$ and each point is assigned to a grid cell based on its coordinates. For a given point $i \in \mathcal{G}_i$, its token shift feature is defined as:
\begin{equation}
\begin{aligned}
\label{eq-a-2}
\hat{f_i} &=\sum_{j \in \mathcal{G}_i} w_{ij}f_j ,
\quad w_{ij} &= {\frac{\exp(-\|x_j - \mu_i\|)}{\sum_{k \in \mathcal{G}_i} \exp(-\|x_k - \mu_i\|)}} ,
\end{aligned}
\end{equation}

\noindent where $ \mu_i = \frac{1}{|\mathcal{G}_i|} \sum_{j \in \mathcal{G}_i} x_j $ denotes the geometric center of subregion $ \mathcal{G}_i $, and $ w_{ij} $ is the contribution of point $j$ to the representation of point $i$, with higher weights assigned to closer points. 
To preserve the discriminability of the original features and avoid excessive perturbation,
we selectively perturb only a subset of channels and introduce a residual fusion mechanism to ensure stable feature refinement:
\begin{equation}
\begin{aligned}
\label{eq-a-3}
f_i^{\text{out}} = [\lambda f_i^{(1)} + (1 - \lambda) \hat{f}_i^{(1)} \; \| \; f_i^{(2)}],
\end{aligned}
\end{equation}
where $ f_i^{(1)} $ represents the first $C^{'}$ channels (used for disturbance), $ f_i^{(2)} $  is the remaining channel, and $\lambda\in (0, 1) $ controls the degree of the disturbance in token shift.

\noindent \textbf{Remark.} Note that our presented AGT-Shift module does not rely on KNN or explicit adjacency graph construction, nor does it depend on additional parameter learning. All aggregation processes can be completed through tensor operations, and the overall computational complexity is $ \mathcal{O}(N)$.

\begin{table*}[t]
    \renewcommand\arraystretch{1.15} 
		\setlength{\tabcolsep}{0.40mm}

		\begin{center}
  \resizebox{1.0\textwidth}{!}{%
		  \begin{tabular}
{c|c| c|c| c c c >{\columncolor{lightgray}}c |cccc>{\columncolor{lightgray}}c}

\hline

\multirow{2}{*}{\bf{Method}} & \multirow{2}{*}{\bf{Setting}} & \multirow{2}{*}{\bf{Venue}} & \multirow{2}{*}{\bf{Backbone}} & \multicolumn{4}{c|}{ \textit{PointDA-10 Benchmark}} & \multicolumn{5}{c}{\textit{PointDG-3to1 Benchmark}} \\

 & & & & \small{M,S*→S} & \small{M,S→S*} & \small{S,S*→M} & \bf{Avg.} & \small{ABC→D} & \small{ABD→C} & \small{ACD→B} & \small{BCD→A} & \bf{Avg.} \\

\hline
PointDAN~\cite{qin2019pointdan} & DA & NeurIPS'2019 & PointNet & 77.38 & 40.32 & 78.69 & 65.46 & 58.85 & 81.66 & 48.86 & 79.95 & 67.33  \\
DefRec~\cite{achituve2021self} & DA & WACV'2021 & DGCNN & 77.23 & 44.28 & 84.77 & 68.76 & 72.76 & 79.97 & 43.29 & 87.94 & 70.99  \\
GAST~\cite{zou2021geometry} & DA & ICCV'2021 &  DGCNN & 79.43 & 47.69 & 81.72 & 69.61 & 71.78 & 86.43 & 52.31 & 86.21 & 74.18  \\
\hline
MetaSets~\cite{huang2021metasets} & DG  & CVPR'2021 & PointNet & 81.39 & 50.86 & 83.48 & 71.91 & 73.24 & 92.41 & 60.97 & 87.28 & 78.48  \\
PDG~\cite{wei2022learning} & DG & NeurIPS'2022 &  PointNet & 79.82 & 51.73 & 83.51 & 71.69 & 73.38 & 92.98 & 60.57 & 89.90 & 79.21  \\
PointNeXt~\cite{qian2022pointnext} & DG & NeurIPS'2022 & PointNet & 77.31 & 43.32 & 78.16 & 66.26 & 71.47 & 91.70 & 46.39 & 88.95 & 74.63 \\

X-3D~\cite{sun2024x} & DG & CVPR'2024 & PointNet & 78.06 & 46.91 & 79.69 & 68.22 & 71.58 & 91.89 & 48.34 & 88.45 & 75.07 \\

PCT~\cite{guo2021pct} & DG  & CVM'2021 & PointTrans & 80.23 & 48.29 & 81.91 & 70.14 & 71.43 & 87.43 & 58.43 & 88.34 & 76.41  \\

GBNet~\cite{qiu2021geometric} & DG & TMM'2021 & PointTrans & 79.94 & 48.92 & 81.34 & 70.07 & 72.78 & 87.83 & 57.76 & 88.82 & 76.80  \\

SUG~\cite{huang2023sug} & DG & MM'2023 &  PointTrans & 78.34 & 49.59 & 82.03 & 69.99 & 71.58 & 89.62 & 54.66 & 86.35 & 75.55  \\

PCM~\cite{guo2021pct} & DG  & AAAI'2025 & PCM & 81.02 & 46.83 & 83.92 & 70.59 & 72.27 & 91.24 & 57.28 & 87.54 & 77.08 \\

PointDGMamba~\cite{yang2025pointdgmamba} & DG  & AAAI'2025 & PCM & 84.33 & 52.83 & 87.38 & 74.85 & 74.20 & 95.51 & 61.71 & 90.68 & 80.53\\

V-RWKV'~\cite{duan2024vision} & DG  & ICLR'2025 & V-RWKV & 81.90 & 49.52 & 85.49 & 72.24 & 73.42 & 92.12 & 57.88 & 88.18 & 77.90\\

\hline
PointDGRWKV & DG  & - & V-RWKV & \textbf{84.39} & \textbf{54.10} & \textbf{88.49} & \textbf{75.66} & \textbf{76.37} & \textbf{95.99} & \textbf{63.92} & \textbf{91.38} & \textbf{81.92} \\
\hline

            \end{tabular}
            }
        \end{center}
\vspace{-3mm}
\caption{Performance comparison between the proposed method and the state-of-the-art point cloud classification methods on the PointDA-10 and PointDG-3to1 benchmarks. The metric used is overall classification accuracy (\%), and \textbf{Avg.} indicates the mean accuracy across all target domain scenarios. The highest result in each benchmark is marked in \textbf{bold}.}
\label{tab-first}
\end{table*}

\subsection{Cross-Domain Key Distribution Alignment}

We observe that there are significant differences in the distribution of $\mathbf{k}$ across different domains, such as high or low mean values of $\mathbf{k}$ features and significant differences in variance in some domains. These distribution shifts will cause significant bias at $e^k$ level, leading to shifts of attention focus position within the domain, severely undermining the model's generalizability to unseen domains.

To address this issue, we propose Cross-Domain \textit{Key} Feature Distribution Alignment(CD-KDA) to enhance the modeling stability and structural generalization ability of attention mechanisms in cross-domain scenes.
Regarding the nature of point cloud data, the \textit{key} vector $\mathbf{k}$ 
encodes the relative importance of each point within its local neighborhood or in relation to the global context. Specifically, $\mathbf{k}$  captures the spatial selection tendencies of the points: its mean $\mu$ reflects the global attention focus, while its variance $\Sigma$ characterizes the semantic or geometric diversity among points. Consequently, \textit{if different source domains exhibit distinct distributions of $\mathbf{k}$  due to geometric discrepancies, it will lead to domain shifts in the aggregation of point cloud structures governed by attention mechanisms.}
In contrast, although the \textit{value} vector  $\mathbf{v}$ also contributes to attention computation, it only serves as the weighted content to be aggregated. It does not influence the generation of attention weights directly, nor does it appear within the exponential function of the attention formulation. As such, its effect on generalization is less immediate and critical than that of $\mathbf{k}$.

Additionally, the spatial decay parameter 
$\mathbf{w}$ and bias $\mathbf{u}$ in Bi-WKV~\cite{duan2024vision} are shared model parameters that encode the model’s inherent sensitivity to spatial distance and positional priors. These parameters should be learned jointly across multiple source domains to capture a unified inductive bias, and therefore should not be forcibly aligned across domains.

Based on these observations, we argue that \textit{aligning the dynamic input key representations 
$\mathbf{k}$ across source domains is the most critical and effective strategy to enhance generalization.}  Let the set of source domains be denoted as $\{\mathcal{D}_1, \mathcal{D}_2, \dots, \mathcal{D}_n\}$, with the corresponding key features extracted from each domain represented by $\mathbf{k}^{(i)} \in \mathbb{R}^{T \times C}$. We define the following objective function for alignment:
\begin{equation}
\begin{aligned}
\label{eq-a-3}
\mathcal{L}_{\text{CD-KDA}} = \frac{1}{|\mathcal{P}|} \sum_{(i, j) \in \mathcal{P}} \left\| \mu^{(i)} - \mu^{(j)} \right\|_2^2 + \left\| \Sigma^{(i)} - \Sigma^{(j)} \right\|_F^2,
\end{aligned}
\end{equation}
where $\mu^{(i)} = \frac{1}{T} \sum_t \mathbf{k}^{(i)}_t$ is the \textit{key} mean of the i$\text{-}$th domain, $\Sigma^{(i)}$ is the corresponding covariance matrix, $\mathcal{P}$ is the set of unordered domain pairs between all source domains, and $\| \cdot \|_F$ is the Frobenius norm. 
As such, by minimizing the distribution of $\mathbf{k}$ representations between source domains, the cross-domain stability of the attention mechanism is enhanced. The introduced CD-KDA explicitly aligns the distribution of $\mathbf{k}$ representations in different source domains on a source domain basis, thereby improving the consistency of the model's attention distribution to the unseen domain.

\subsection{Training and Inference}

During the training phase, the total loss of the model includes the classification loss and the cross-domain feature alignment loss, as follows:
\begin{equation}
\begin{aligned}
\label{eq-E-1}
\mathcal{L} = \lambda_1 \mathcal{L}_{\text{cls}} + \lambda_2 \mathcal{L}_{\text{CD-KDA}},
\end{aligned}
\end{equation}
where $\mathcal{L}\_{\text{cls}}$ is the cross-entropy loss used to supervise the model’s ability to recognize point cloud classes. Hyperparameters $\lambda_1$ and $\lambda_2$, respectively, adjust the weights of these two losses during training. 
During the inference stage, only the trained feature extractor and classifier are used. The model no longer requires access to any source domain data and directly predicts the class labels of the target domain samples. This allows for efficient and scalable deployment in unseen domains without additional adaptation.

\section{Experiments}
\subsection{Experiments Setup}
\label{experiment1}

{\bf{Implementation Details.}} Our model was trained on an NVIDIA RTX 4090 GPU. The optimizer AdamW \cite{loshchilov2017decoupled} is used, with an initial learning rate of $1\times 10^{-4} $, a cosine annealing scheduling strategy, and a weight decay of $1\times10^{-4}$. 
During the training process, preprocessing and enhancement operations such as scaling, normalization, and random jitter were applied to the input point cloud data. The model adopts a four-stage hierarchical structure for gradually extracting and aggregating multi-scale point cloud features. The number of RWKV blocks included in each stage is 1, 1, 2 and 2, respectively. In all experiments, 
$\lambda_1 \text{ = } 1$ and $\lambda_2 \text{ = } 0.3 $ by default.

\noindent {\bf{Benchmarks.}} To evaluate the generalizability of our method in DG PCC, we conduct experiments on two benchmarks. The first is PointDA-10~\cite{qin2019pointdan, dai2017scannet, wu20153d}, which includes ModelNet-10 (M), ShapeNet-10 (S), and ScanNet-10 (S*) with 10 shared categories. ModelNet and ShapeNet contain clean point clouds generated from synthetic 3D models, while ScanNet captures real-world scenes with frequent missing regions due to occlusion. It defines three cross-domain settings: M, S*→S; M, S→S*; and S, S*→M. The second benchmark is PointDG-3to1~\cite{yang2025pointdgmamba}, including ModelNet-5 (A), ScanNet-5 (B), ShapeNet-5 (C), and 3D-FUTURE-Completion (D)~\cite{liu2024cloudmix,fu20213d}, sharing five classes. It adopts a “leave-one-out” setting to form four settings: ABC→D, ABD→C, ACD→B, and BCD→A. 
Following common DG practices, the training uses only source samples, and the evaluation is performed on the target domain's testing set.

\subsection{Comparison Results}

\noindent {\bf{Comparison Methods.}} 
To comprehensively evaluate the proposed method, we compare it with representative point cloud classification models, including CNN-based methods such as PointDAN~\cite{qin2019pointdan}, DefRec~\cite{achituve2021self}, GAST~\cite{zou2021geometry}, PDG~\cite{wei2022learning}, MetaSets~\cite{huang2021metasets}, PointNeXt~\cite{qian2022pointnext}, and X-3D~\cite{sun2024x}, as well as Transformer-based approaches like SUG~\cite{huang2023sug}, PCT~\cite{guo2021pct}, and GBNet~\cite{qiu2021geometric}. We also include Mamba-based methods PCM~\cite{zhang2025point} and PointDGMamba~\cite{yang2025pointdgmamba}. Additionally, we evaluate V-RWKV, a modified Vision-RWKV~\cite{duan2024vision} variant with a different number of blocks. 
Due to the unavailability of training code, PointRWKV~\cite{he2025pointrwkv} is excluded.

\noindent {\bf{Benchmark Results.}} 
We conduct a comprehensive evaluation of the proposed PointDGRWKV method on two widely used multi-domain point cloud generalization benchmarks: PointDA-10 and PointDG-3to1. The results are presented in Table~\ref{tab-first}. PointDGRWKV consistently outperforms existing methods in terms of average overall accuracy across both benchmarks. Specifically, on the three domain generalization tasks of PointDA-10, PointDGRWKV achieves better performance than the state-of-the-art PointDGMamba across all DG tasks, with an average accuracy of 75.66\%. Notably, on the PointDG-3to1 benchmark, PointDGRWKV achieves an average accuracy of 81.92\% across four domain shifts, outperforming PointDGMamba by a significant margin of 1.39\%. The improvement is particularly pronounced in the most challenging ACD→B setting.

\noindent {\bf{Analysis of Improvements.}}
Overall, PointDGRWKV demonstrates high performance across different domains, indicating strong cross-domain stability. We attribute this consistent improvement to the AGT-Shift mechanism, which better captures the local geometric structures inherent to unstructured point cloud data, effectively mitigating the information mismatch caused by the “pseudo-local receptive field” issue in the original RWKV. Additionally, the CD-KDA module alleviates attention misalignment in Bi-WKV caused by domain-specific variations in \textit{key} distributions, enabling the model to learn more consistent structural perception across source domains and thereby enhancing generalization to unseen domains. It is worth noting that while the proposed method shows clear improvements on average, its advantage is relatively modest in certain simpler settings, suggesting that the primary gains come from improved robustness and stability under more complex domain shifts.

\begin{table}[t]
    \renewcommand\arraystretch{1.15} 
		\setlength{\tabcolsep}{0.85mm}
		\begin{center}
			\begin{tabular}
{c| c| ccc|  >{\columncolor{lightgray}} c >{\columncolor{lightgray}} c}

\hline
\bf{AGTS} & \bf{KDA} & M,S*→S & M,S→S* & S,S*→M & \bf{Avg.} & \bf{Gain} \\
				
\hline
& & 81.70 & 49.52 & 85.49 & 72.24 & - \\
\checkmark & & 83.39 & 51.10 & 86.21 & 73.57 & 1.33\\
& \checkmark & 82.50 & 53.48 & 86.86 & 74.28 & 2.04\\
    
\hline
\checkmark & \checkmark & 84.39 & 54.10 & 88.49 & 75.66 & 3.42\\

\hline
\end{tabular}
\end{center}

\vspace{-3mm}
\caption{Ablation study on the AGT-Shift (AGTS) and CD-KDA (KDA) modules on the PointDA-10 benchmark.}
    \label{tab2}
\end{table}

\begin{table}[t]

    \renewcommand\arraystretch{1.15} 
		\setlength{\tabcolsep}{1.2mm}
		\begin{center}
			\begin{tabular}
{c| ccc|  >{\columncolor{lightgray}} c}

\hline
 \bf{Shift} & M,S*→S & M,S→S* & S,S*→M & \bf{Avg.}  \\

\hline
KNN-RandOne         & 83.39 & 52.85 & 87.38 & 74.54 \\
KNN-Avg             & 83.87 & 53.36 & 85.51 & 74.25 \\
KNN-WAvg            & 83.35 & 53.31 & 86.80 & 74.82 \\

\hline
AGT-Shift (Ours)    & 84.39 & 54.10 & 88.49 & 75.66\\

\hline
\end{tabular}
\end{center}
\vspace{-3mm}
\caption{Ablations on different shifting strategies on the PointDA-10 benchmark.}
\vspace{-5mm}
    \label{ablation-shift}

\end{table}

\subsection{Ablation Study}
\noindent {\bf{Effectiveness of AGT-Shift and CD-KDA.}} 
To further validate the specific roles of the proposed components in the model, we conducted ablation experiments on the PointDA-10 benchmark, and the results are presented in Table~\ref{tab2}. Firstly, we constructed a basic version of V-RWKV' without AGT-Shift and CD-KDA modules. Compared with the basic model, the introduction of AGT-Shift improved the overall performance in all three tasks, indicating that this module has a positive effect on modeling the local geometric structure of point clouds. Furthermore, we separately introduced the CD-KDA module for evaluation. We observed stable performance improvements, indicating that the cross-domain \textit{key} feature alignment mechanism has a certain effect in alleviating attention bias and improving generalization ability. Finally, when both modules are introduced simultaneously, the model achieves better performance on all transfer tasks, indicating that AGT-Shift and CD-KDA are complementary, jointly promoting the overall performance.

\noindent {\bf{Effects of Different Shifting Strategies.}}
To evaluate the effectiveness of our proposed AGT-Shift module, we compareit with three different token shifting strategies:
(1) KNN-Random Replacement (KNN-RandOne): For each point, its K nearest neighbors are first identified using KNN search. Then, one neighbor is randomly selected to replace the original point’s feature. 
(2) KNN-Mean Aggregation (KNN-Avg): After obtaining the K nearest neighbors, the output feature is computed as the average of all neighbor features, replacing the original.
(3) KNN-Weighted Aggregation (KNN-WAvg): Different from KNN-Avg, a soft weighting scheme is applied based on spatial distance, where closer neighbors contribute more.
Table~\ref{ablation-shift} shows that the performances of these strategies are
74.54\%, 74.25\%, 74.82\% for KNN-RandOne, KNN-Avg, KNN-WAvg, respectively.
Notably, KNN-RandOne performs competitively despite its simplicity, suggesting that introducing randomness can help alleviate overfitting to local patterns. However, all three variants suffer from quadratic computational complexity due to KNN. In contrast, our AGT-Shift achieves better performance while maintaining linear complexity and avoiding pairwise distance computation, highlighting its efficiency and robustness in large-scale domain generalization tasks.

\noindent {\bf{Impact of \textit{Key} and \textit{Value} Alignment in CD-KDA.}}
To further investigate the roles of different components in the attention mechanism, we conduct an ablation study isolating the effects of the \textit{key} ($\mathbf{k}$) and \textit{value} ($\mathbf{v}$) features in our proposed CD-KDA module. Specifically, we design the following variants:
(1) None: no alignment is performed, serving as a baseline;
(2) Only $\mathbf{v}$: alignment is applied solely on the $\mathbf{v}$ features;
(3) $\mathbf{k}$ and $\mathbf{v}$: both $\mathbf{k}$ and $\mathbf{v}$ are aligned simultaneously;
(4) Only $\mathbf{k}$ (Ours): alignment is applied solely on the \textit{key} representations $\mathbf{k}$, as proposed in our method.
The results in Table~\ref{ablation-kv} show that aligning only the value vector $\mathbf{v}$ brings limited performance improvement, suggesting that despite contributing to feature aggregation, $\mathbf{v}$ has a relatively minor influence on cross-domain generalization. In contrast, aligning both the key $\mathbf{k}$ and value $\mathbf{v}$ vectors leads to a more noticeable performance gain, indicating that promoting feature consistency does benefit generalization. Interestingly, the best performance is achieved when alignment is applied solely to $\mathbf{k}$, confirming our hypothesis that the key vector, which directly influences attention weights via the exponential function, plays a more critical role in guiding spatial focus and structural understanding. Thus, aligning $\mathbf{k}$ across domains significantly stabilizes the attention mechanism and enhances generalization performance.

\subsection{Visualization and Analysis}

\begin{table}[t]
    \renewcommand\arraystretch{1.15} 
		\setlength{\tabcolsep}{1.5mm}
		\begin{center}
			\begin{tabular}
{c| ccc|  >{\columncolor{lightgray}} c}

\hline
 \bf{Setting} & M,S*→S & M,S→S* & S,S*→M & \bf{Avg.}  \\
				
\hline
None                            & 83.39 & 51.10 & 86.21 & 73.57 \\
Only $\mathbf{v}$               & 83.67 & 51.89 & 86.68 & 74.08 \\
$\mathbf{k}$ and $\mathbf{v}$   & 84.63 & 54.07 & 88.34 & 75.68 \\
    
\hline
Only $\mathbf{k}$ (Ours)        & 84.39 & 54.10 & 88.49 & 75.66\\

\hline
\end{tabular}
\end{center}
\vspace{-5mm}
\caption{Ablation study comparing different alignment settings for key ($\mathbf{k}$) and value ($\mathbf{v}$) in the CD-KDA module.}
    \label{ablation-kv}
\end{table}

\begin{figure}[t]
\centering
\includegraphics[scale=0.55]{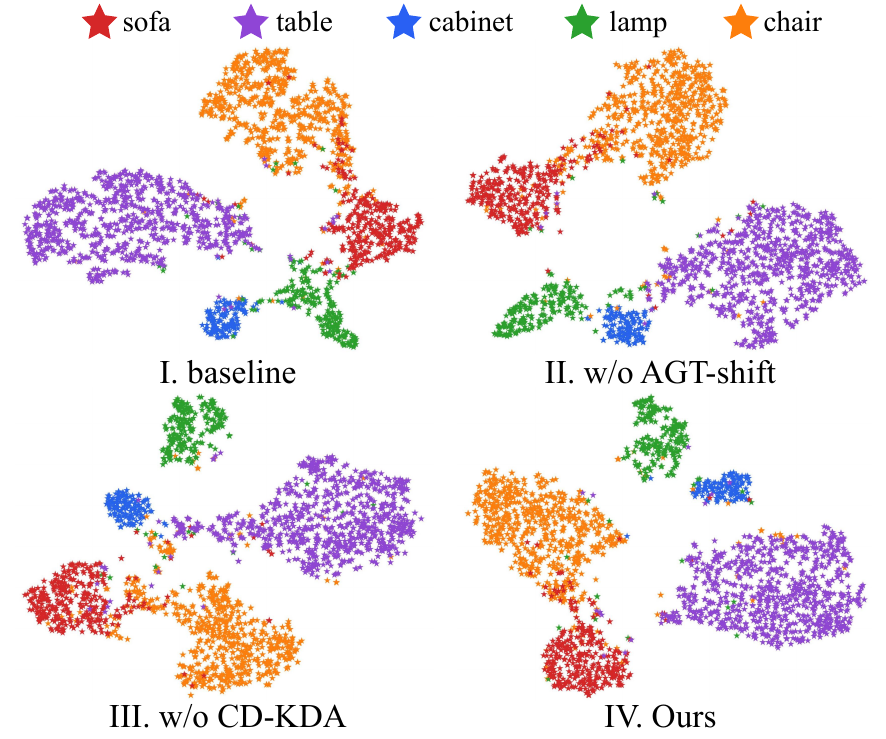}
\caption{T-SNE visualization results of target domain feature distribution of PointDGRWKV under different module configurations. Different colors represent different classes.}
\label{5.visualization}
\vspace{-3mm}
\end{figure}

\noindent {\bf{T-SNE Feature Visualization.}}
To investigate the effectiveness of each proposed module, we visualize the features of target distributions under four different configurations using t-SNE, as illustrated in Fig.~\ref{5.visualization}. Specifically, (I) shows the baseline without AGT-Shift and CD-KDA, (II) removes only the AGT-Shift module, and (III) removes only the CD-KDA module, while (IV) represents our complete model. The visualization is conducted on the test set of the ShapeNet-5 (C) dataset under the PointDG-3to1 benchmark, with different colors denoting different classes.
Among the first three ones, the baseline (I) exhibits the lowest intra-class compactness, indicating limited discriminability. Removing either AGT-Shift (II) or CD-KDA (III) leads to moderate improvements, but both still show less compact clusters and more ambiguous boundaries compared to the full model. Notably, these differences are especially clear in categories such as cabinet (blue) and table (purple). The complete model (IV), in contrast, achieves the most compact intra-class distributions and the clearest inter-class boundaries, demonstrating superior feature separability and confirming the essential roles of AGT-Shift and CD-KDA in DG.

\begin{table}[t]
    \renewcommand\arraystretch{1.15} 
		\setlength{\tabcolsep}{1.4mm}
\begin{center}

    \begin{tabular}
		{ c|ccc| >{\columncolor{lightgray}}c}

		\hline
        \bf{Scale} & M,S*→S & M,S→S* & S,S*→M & \bf{Avg.} \\
        \hline
        Ours-Base   & 83.99 & 53.83 & 87.62 & 75.15 \\
        Ours-Standard  & 84.39 & 54.10 & 88.49 & 75.66\\
        Ours-Large  & 84.63 & 54.61 & 89.14 & 76.13 \\
        \hline
        
    \end{tabular}
\end{center}

\vspace{-4mm}
\caption{Generalization results of PointDGRWKV across varying network scales.}
    \label{tab-scale}
\end{table}

\begin{table}[t]
    \renewcommand\arraystretch{1.05} 
		\setlength{\tabcolsep}{1.8mm}

		\begin{center}
          \resizebox{0.48\textwidth}{!}{%
		  \begin{tabular}
{c |ccc| >{\columncolor{lightgray}}c }
\hline
 & Params & GFlops & Time  &  \\
\multirow{-2}{*}{\bf{Method}} & (M) & (G) & (ms) & \multirow{-2}{*}{\bf{Acc(\%)}}\\
 \hline
GAST~\cite{zou2021geometry} & 75.36 & 2.17 & 23.13 & 69.61 \\
PCT~\cite{guo2021pct} & 2.88 & 2.19 & 27.65 & 70.14 \\
GBNet~\cite{qiu2021geometric} & 8.77 & 9.87 & 80.97 & 70.07 \\
SUG~\cite{huang2023sug} & 19.17 & 18.4 & 5.42 & 69.99 \\
PCM~\cite{guo2021pct} & 35.85 & 20.18 & 6.26 & 70.59 \\
PointDGMamba~\cite{yang2025pointdgmamba} & 13.09 & 6.08 & 3.35 & 74.85 \\

\hline

Ours-Base   & 2.13  & 3.22 & 1.68 & 75.15 \\
Ours-Standard  & 3.72  & 4.57 & 2.39 & 75.66 \\
Ours-Large  & 10.40 & 7.60 & 2.92 & 76.13 \\

\hline

            \end{tabular}
            }
        \end{center}
\vspace{-5mm}
\caption{Analysis of Computational efficiency on the single NVIDIA 4090 GPU.}
\vspace{-5mm}
\label{tab-cost}
\end{table}

\noindent {\bf{Effect of Model Scale.}} To examine the impact of model capacity on generalization, we design three variants of our PointDGMamba: Ours-Base, Ours-Standard, and Ours-Large. As summarized in Table~\ref{tab-scale}, the Standard version follows the default configuration used in our main experiments. Ours-Base reduces the number of network blocks by half, resulting in a shallower architecture. In contrast, Ours-Large increases the feature dimension and performs denser point sampling during processing. Among these, Ours-Large achieves the best average accuracy, indicating that increased representational power and finer geometric granularity benefit generalization. Meanwhile, the Base model still performs competitively to state-of-the-art methods, indicating the proposed method is effective even at lower computational cost.

\noindent {\bf{Analysis of Computational Efficiency.}}
To assess the efficiency of our approach, we report comparisons in terms of model parameters, GFlops, inference time, and accuracy, as summarized in Table~\ref{tab-cost}. Our method delivers strong generalizability while maintaining lower computational cost than most existing methods. Notably, Ours-Base variant achieves 75.15\% accuracy with only 2.13M parameters, 3.22 GFlops, and 1.68 ms inference time, highlighting the effectiveness of our lightweight design. These results confirm that even the compact version of our model can achieve competitive performance with minimal cost, demonstrating strong potential for deployment in resource-constrained scenarios.

\section{Conclusion}
We propose PointDGRWKV, the first RWKV-based framework for domain generalization in point cloud classification. It enhances spatial perception and generalization while retaining RWKV’s efficient sequence modeling and linear complexity. To address RWKV’s limitations on 3D data, we introduce AGT-Shift for improved local geometric modeling and CD-KDA to reduce attention drift by aligning key distributions across domains. Extensive experiments on PointDA-10 and PointDG-3to1 benchmarks confirm that our method achieves state-of-the-art performance with a strong balance between efficiency and robustness.

{
    \small
    \bibliographystyle{ieeenat_fullname}
    \bibliography{main}
}

\end{document}